\documentclass[11pt,a4paper]{article}

\usepackage[nohyperref]{emnlp2020}
\usepackage{times}
\usepackage{latexsym}
\usepackage{amsmath}
\usepackage{graphicx}
\usepackage{hyperref}
\usepackage{adjustbox}
\usepackage{microtype}
\aclfinalcopy 
\title{WER we are and WER we think we are}

\author{Piotr Szymański$^{1,2,*}$, Piotr Żelasko$^{3,*}$, Mikołaj Morzy$^{1,4}$, Adrian Szymczak$^{1}$, \\ \textbf{Marzena Żyła-Hoppe$^{1}$, Joanna Banaszczak$^{1}$, Łukasz Augustyniak$^{1,3}$}\\ \textbf{Jan Mizgajski$^{1,4}$, Yishay Carmiel$^{1}$} \\ $^{1}$ Avaya Inc., USA \\ $^{2}$ Wrocław University of Science and Technology, Wrocław, Poland \\ $^{3}$ Center for Language and Speech Processing, Johns Hopkins University, Baltimore, MD, USA \\ $^{4}$ Poznań University of Technology, Poznań, Poland\\ $^{*}$ Equal contribution   \\   {\tt \href{mailto:piotr.szymanski@pwr.edu.pl}{piotr.szymanski@pwr.edu.pl}}, {\tt \href{mailto:pzelasko@jhu.edu}{pzelasko@jhu.edu} }}

\date{\today}

\begin{document}
\maketitle

\begin{abstract}
Natural language processing of conversational speech requires the availability of high-quality transcripts. In this paper, we express our skepticism towards the recent reports of very low Word Error Rates (WERs) achieved by modern Automatic Speech Recognition (ASR) systems on benchmark datasets. We outline several problems with popular benchmarks and compare three state-of-the-art commercial ASR systems on an internal dataset of real-life spontaneous human conversations and HUB'05 public benchmark. We show that WERs are significantly higher than the best reported results. We formulate a set of guidelines which may aid in the creation of real-life, multi-domain datasets with high quality annotations for training and testing of robust ASR systems.
\end{abstract}

\section{Introduction}
\label{sec:intro}


The last few years have witnessed unprecedented progress in Automatic Speech Recognition (ASR) systems. They have become ubiquitous in everyday lives, from phones voice assistants through dictation of text messages and e-mails to managing household appliances with home assistants. It is not surprising that major vendors are trying to showcase the quality and accuracy of their products. A comprehensive benchmark of available ASRs~\cite{synnaeve_2020} cites word error rates (WERs) as low as 2\%–3\% on standard datasets. These reports may incur a false conviction that automatic speech recognition is mostly a solved problem. Nothing could be further from the truth.

One possible cause for this misconception and gross over-estimation of the accuracy of contemporary ASRs is the confounding of two regimes of speech recognition. The majority of interactions with ASRs happen in the context of chatbot-like interactions, when a human is fully aware of speaking to a machine. In these circumstances most people simplify their utterances, speaking in short, well-structured phrases which obey the correct grammatical structure of either an interrogative or an imperative sentence. \citet{siegert2018we} point out that conversations between people are more lively and dynamic, while the conversations with Alexa remain mostly simple request/response interactions. Significant acoustic differences allowed them to build a clasifier which achieves 81\% of accuracy for distinguising between human-human and human-chatbot interactions. Human-chatbot interactions are in stark contrast with typical spontaneous human-human conversations. Such conversations are riddled with various disfluencies (discourse markers, filled pauses, back-channeling), lack clear phrase borders, and human utterances are often not terminated correctly. Oftentimes information is exchanged using non-verbal means, for example, prosody, paralanguage, and non-verbal vocalizations. \citet{hill2015real} found that human–chatbot communication lacked much of the richness of vocabulary found in conversations among people. All these factors make the correct recognition of spontaneous human conversations a very challenging task.


We do not share the optimistic view of the overall ASR accuracy. In our opinion, ASR systems still have a long way to go toward robust recognition of spontaneous human conversations. Popular ASR benchmarks, such as Librispeech~\cite{panayotov2015librispeech}, WSJ~\cite{paul1992design}, Callhome, Fisher~\cite{cieri2004fisher}, or Switchboard~\cite{godfrey1992switchboard}, are -- to a different extent -- misaligned with the contemporary domains of ASR applications. Some of these benchmarks are too simple to truly challenge ASRs, while the more challenging ones do not cover the full spectrum of inputs encountered in everyday operations. As such, these benchmarks do not provide reliable estimations of the actual ASR accuracy. On the other hand, on the truly challenging benchmarks, such as the dinner party conversations CHiME5~\cite{manohar2019acoustic} benchmark, modern ASRs report WERs in the ranges of 46\%–73\%.


Another problem is the domain adaptation. Both Fisher and Switchboard corpora, although trying to mimic real, spontaneous conversations, are inherently artificial. The protocols for the creation of both corpora involved pairs of voice actors having a conversation on a random subject drawn from a collection of predefined topics. These conversations are very different from real-life spontaneous conversations, where the topics may vary greatly, but at the same time the domain of application imposes strict constraints on the vocabulary and the form of the conversations. Many benchmark datasets represent scripted or semi-scripted conversations (EU Parliament speech transcripts, TED talk transcripts). There are consequential differences between scripted and spontaneous conversations and they affect the results of the ASR evaluation~\cite{Shriberg2001}.

\begin{figure*}[t]
\centering
\includegraphics[width=0.8\linewidth]{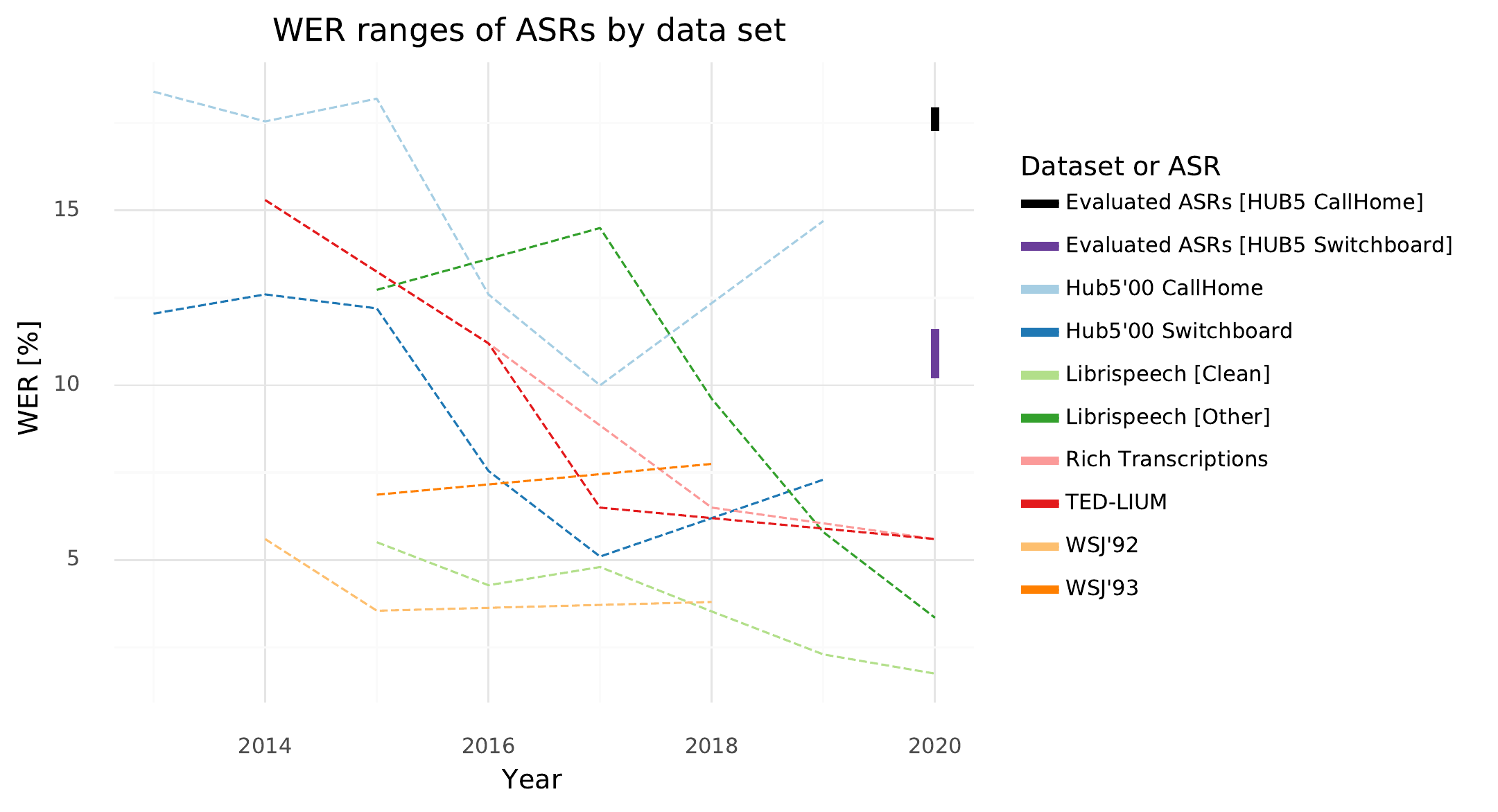}
\vspace{-2pt}
\caption{WER ranges in ASR systems. Reference values taken from the \textit{Wer are we} report \cite{synnaeve_2020} and the \textit{Papers with Code} website \cite{papers_with_code} for ASR solutions published in the last 5 years. Outliers were removed for the sake of figure readability.\label{fig:wers}}
\centering
\vspace{-5pt}
\end{figure*}

Finally, benchmark datasets tend to be quite homogeneous with respect to the demographic features of voice actors. In real-life conversations, factors such as age, gender, ethnicity, and dialect are important determinants of speech particularities. Non-native language speakers are virtually absent from benchmark datasets, and the novelties introduced by these speakers, both acoustic (pronunciation) and linguistic (vocabulary and syntax), are not accounted for in the results of ASR evaluation~\cite{Koenecke7684}. Even for gender, a variable ostensibly easy to control, we find a huge imbalance in the distribution of the number of utterances in benchmark datasets. Benchmark datasets do not represent the true diversity of real-world conversations, both at input signal characteristics and conversation semantics levels.


In our opinion, what is commonly assumed to constitute the state-of-the-art in the ASR quality is grossly over-estimated. IRL-WERs (in-real-life WERs) are much higher than reported. In this paper, we compile a benchmark of real-life phone call spontaneous conversations from five different domains and present the unbiased comparison of three major cloud ASR systems. We show that the WERs observed both in the public HUB'05 benchmarks and in real conversations are much higher than usually reported, and this phenomenon is observed across all of the tested domains. Our benchmark consists of call center conversations containing almost 3000 utterances. The conversations were transcribed by two professional human annotators. The conclusions are clear: we are definitely not where we think we are in terms of WERs.

The contributions of our paper are twofold. Firstly, we experimentally compare WERs of three major commercial ASR systems on publicly available benchmarks and we contrast these results with our internal benchmark of real-life human spontaneous conversations. Secondly, we raise community awareness regarding the problems caused by the optimistic bias towards the ASR accuracy. We issue a call to action with the aim of addressing this bias by researchers and funding institutions.

\section{The State of the Error Rate}
\label{sec:ser}

We evaluate several ASR systems on a multi-domain dataset of 50 call center conversations recorded at 8~kHZ, using standard, modern telephony quality. The dataset comprises 8.5 hours of audio, including 2.2 hours of speech. These calls consist of 1595 agent and 1361 customer utterances. An average utterance consists of 10 words, thus resembling a sentence-length phrase. For speech recognition we use three different state-of-the-art commercial ASR solutions which we find representative in terms of what is available on the market. When a given ASR vendor offered such an option, we used a telephonic speech model. We report the WER of evaluated ASRs on our dataset and on the HUB5'00 Switchboard and CallHome evaluation subset to allow comparison to the state of the art on publicly available data. 

We collected per channel audio data as RIFF (little-endian) data, WAVE audio, Microsoft PCM, 16 bit, mono 8000 Hz. These audio data files were transcribed by each of the ASR systems and the transcription results were exported as CTM files. We used the NIST Scoring Toolkit (sctk) scripts to obtain WER scores from the reference STM and predicted CTM files, using the established hubscr.pl script. We collected these sctk-based scores into data frame and organized the results into tables per domain and the user of a given channel.

We would wholeheartedly like to provide a more reproducible contribution, however, due to legal constraints, we are required to make the experimental procedure more obscure than we would like. While we understand the scientific expectation of complete experimental transparency and reproducibility, we are not able to provide the community with neither the benchmark data nor the detailed information about evaluated systems.

To set our benchmark in context, we present the median ASR WER per year reported in scientific papers as aggregated by the wer\_we\_are report \cite{synnaeve_2020}. They provide a basis for an optimistic view on both the state of the art in ASRs published by scientific teams and a visible trend of improvement. Depending on the dataset, the error rates, as of 2019, do not exceed 15\%.  Depending on the data origin and quality, WER drops as low as 2\%–4\% for the Librispeech \cite{panayotov2015librispeech} corpus of approximately 1000 hours of 16kHz recordings of English audiobooks. The WSJ'92 and '93 corpora \cite{linguistic1994csr} contain 73 hours of clean speech dictation and clean conversational speech of journalists; and ASRs have admitted a WER of 3\%–7\% on this dataset. The TED-LIUM ASR \cite{rousseau2012ted} task consists of 118 hours of speech recorded in high quality from TED Talks and its WER is reported at 5\%. The currently best WER for the HUB'05\footnote{LDC2002T43} evaluation is at 5\% for the Switchboard part and 9\% for the CallHome part – both of which are phone conversational datasets.

Unfortunately, as we see in Table~\ref{tab:wers-public}, the commercial ASR systems in our evaluation achieve nearly double the error rates on both HUB'05 evaluation subsets. This result may be explained as follows. Firstly, the results are typically reported using the oracle segmentation of the evaluation data, running the speech recognizer on each segment of speech separately. Instead, we evaluated each ASR by providing the whole 5 minutes chunks of audio, so that each system had to perform speech activity detection (SAD) first. This resembles the real-life application usage. Secondly, for each public dataset, the ASR system usually uses a language model (LM) estimated on the training set, making it representative of the domain data. In contrast, commercial ASR systems have to use more general LMs which are likely to perform worse on any specific benchmark. 

\begin{table}[h!]
\centering
 \begin{tabular}{c c c c} 
 \hline
    ASR & CCC    &  SWBD  &  CallHome \\
    \hline
    ASR 1 & 17.9 & 11.62          & 17.69 \\
    ASR 2 & 19.2 & 11.45          & 18.6 \\
    ASR 3 & 16.5 & 10.2          & 15.85 \\
 \hline
 \end{tabular}
 \caption{WER [\%] comparison on benchmarks \label{tab:wers-public}}
 \vspace{-1pt}
\end{table}

The gap between the state-of-the-art results reported on public benchmarks and real-life is even more visible in the case of multi-domain datasets. In Table~\ref{tab:wers-internal} we present WERs obtained on our internal benchmark. The five domains evaluated in the benchmark include booking touristic reservations, finance, two types of insurance domains, and telecommunication conversations. We can see that insurance \#2 was the easiest of domains for the ASRs, followed by finance. The highest error rates are obtained for booking and wireless telecommunication calls, which can be related to the fact that more of their utterances contain entities related to date and time, money, places, and product or company names. We also include the division of WERs per channel. Per channel differences are not large enough to assume that one of the channels is in a significantly better position for performing downstream NLP tasks. 

\begin{table}[h!]
\centering
 \begin{tabular}{c c c c c} 
 \hline
    {} & ASR 1 & ASR 2 & ASR 3 \\ \hline
    Booking  & 	21.19 &  	22.16 &  	20.95 \\ 
    Finance &  	16.82 &  	18.46 &  	15.83 \\ 
    Insurance 1 &  18.01 &  	20.20 &  	17.84 \\ 
    Insurance 2 &  	15.25 &  	17.11 &  	13.73 \\ 
    Telecomm. &  	19.75 &  	23.31 &  	17.62 \\ 
  \hline
    Agent & 	16.97 	& 	17.83 & 	16.49 \\
    Customer & 	17.87 & 	20.99 & 	16.48 \\
 \hline
\end{tabular}
\caption{Internal benchmark WER [\%]}
\label{tab:wers-internal} 
\vspace{-5pt}
\end{table}

\section{Call to Action}
\label{sec:call-2-act}

The main reason for publishing the results presented in Section~\ref{sec:ser} is to support our call to action. We want to encourage the ASR and NLP communities to pursue research and collaboration to address the shortcomings of current ASR benchmarks. We need to collect and annotate audio datasets that are much better aligned with contemporary application domains of ASR systems, work on extended and more inclusive acoustic models representing a much broader spectrum of dialects, account for technological advances which influence physical properties of processed audio signals, and develop language models for multi-domain conversations.

The situation where most available spontaneous conversation datasets are over 20 years old is both easy to overlook and hard to believe. What is worse, judging by the standards one expects from modern NLP corpora, these datasets lack rich annotations. In many cases, part-of-speech tags and dependency structure are missing, and few datasets have complete dialog act annotations. Hardly any datasets contain named entity annotations beyond very basic NER schemes. 

High quality annotations enable building better language models and can improve ASR quality metrics even in the simplest of scenarios when the ASR was trained with a language model aware of part of speech (POS) annotations. With just such basic annotations \citet{stiefel-vu-2017-enriching} reported a small WER decrease even on data sets without proper human annotation, with POS tag obtained via automatic tagging.

The LDC catalog, the most comprehensive repository of language-related corpora, lists only four phone conversation datasets\footnote{LDC2015S08,LDC2019S06,LDC2013S05,LDC2010S02} collected recently (i.e., datasets which are not based on Fisher, Callhome, or Switchboard). None of these new datasets contain rich annotations, and the datasets are not related to real-world, multi-domain conversations -- precisely the type of application the industry is trying to deploy. For more advanced applications, such as machine translation or machine comprehension, such resources are scarce. 

Recently published datasets for spoken question answering are synthetically produced by text-to-speech systems~\cite{lee2018odsqa,li2018spoken}. This artificial generative procedure strips training data of many subtle characteristics of human speech, such as prosodic features and non-verbal acoustic cues. Many aspects of spontaneous human speech are difficult to model in a text-to-speech synthesizer, rendering the resulting recordings less aligned with data on which ASRs will be applied.

These problems are not insurmountable. A thoughtful collaboration between academia and industry partners can lead to the creation of high-quality training and testing datasets. Importantly, these datasets should -- preferably -- be published under an open-access license, or at least should be made available through openly accessible data platforms like Mozilla's Common Voice. 

It is of important to address the legal issues of open-sourcing the recordings of the human voice, which currently hold back the industry from releasing conversational speech datasets such as ours.
This is an opportunity for funding institutions to spark interdisciplinary research and economic growth through R\&D breakthroughs in the area of automatic speech recognition.

There are multiple factors which may contribute to the progress in ASR for human conversational speech. These include funding schemes, grant programs, data donations, and student internships, to name a few. 

In our opinion, a comprehensive program for advancing the field should encompass the following:

\begin{itemize}
    \item preparing new audio and transcript datasets with rich annotations including: POS tags, dependency structure, entity spans, sentiment annotations, question and answer pairs, dialogue \cite{bunt2011semantics} and discourse annotation \cite{louwerse2003toward}, and augmenting existing corpora with NLP annotations;
    \item developing methods and tools for improving ASR acoustic and language model training and adapting NLP models and pipelines to conversational applications;
    \item developing tools that allow collecting conversational speech and recording real, spontaneous conversations that could be crowdsourced and released openly like Librispeech;
    \item designing open solutions that can serve as common benchmarks for joint ASR+NLP tasks to monitor the progress of the field;
    \item organizing crowd-sourcing collection efforts similar to Mozilla CommonVoice where users could donate their phone calls and/or transcriptions;
    \item constructing new ASR quality measures, based on more richly annotated data, to better evaluate various aspects of transcription quality.
\end{itemize}

\section{Conclusions}
\label{sec:prior}

We argue that contemporary ASR systems cannot cope with spontaneous human conversations satisfactorily, contrary to many beliefs in the NLP community. To substantiate our claim we compile a benchmark consisting of real-world, multi-domain phone call spontaneous conversations on which we observe higher WERs than on traditional datasets. We believe that the overly optimistic perception of ASR accuracy is detrimental to the development of conversational NLP downstream applications. In order to overcome this problem, we ask the community to engage in interdisciplinary research between the academia and industry partners in both ASR and NLP domains and we discuss actions that can be taken. We hope that our call to action will provide some guidelines toward the improvement of ASR systems in the upcoming decade.

\bibliographystyle{acl_natbib}
\bibliography{refs}

\end{document}